\pdfoutput=1

\documentclass[11pt]{article}

\usepackage[]{acl}

\usepackage{times}
\usepackage{latexsym}

\usepackage[T1]{fontenc}

\usepackage[utf8]{inputenc}

\usepackage{microtype}

\usepackage{inconsolata}

\usepackage{graphicx}
\usepackage{booktabs}
\usepackage{makecell}
\usepackage{stfloats}
\usepackage{caption}
\usepackage{subcaption}
\usepackage{amsmath}
\usepackage{url}

\DeclareRobustCommand{\brkbinom}{\genfrac(){0pt}{}}

\title{Visual Cues Enhance Predictive Turn-Taking for Two-Party Human Interaction}

\author{Sam O'Connor Russell \and Naomi Harte \\
        ADAPT Centre, School of Engineering, Trinity College Dublin, Ireland \\ \texttt{\{russelsa,nharte\}}@tcd.ie
        }

\begin{document}
\maketitle
\begin{abstract}
Turn-taking is richly multimodal. Predictive turn-taking models (PTTMs) facilitate naturalistic human-robot interaction, yet most rely solely on speech. We introduce MM-VAP, a multimodal PTTM which combines speech with visual cues including facial expression, head pose and gaze. We find that it outperforms the state-of-the-art audio-only in videoconferencing interactions (84\% vs. 79\% hold/shift prediction accuracy). Unlike prior work which aggregates all holds and shifts, we group by duration of silence between turns. This reveals that through the inclusion of visual features, MM-VAP outperforms a state-of-the-art audio-only turn-taking model across all durations of speaker transitions. We conduct a detailed ablation study, which reveals that facial expression features contribute the most to model performance. Thus, our working hypothesis is that when interlocutors can see one another, visual cues are vital for turn-taking and must therefore be included for accurate turn-taking prediction. We additionally validate the suitability of automatic speech alignment for PTTM training using telephone speech. This work represents the first comprehensive analysis of multimodal PTTMs. We discuss implications for future work and make all code publicly available.
\end{abstract}

\section{Introduction}
\label{sec:intro}

There is 200 ms of silence on average between speaking turns in a
two-party interaction \cite{stivers2009universals,levinson2015timing}, yet language formation takes at least 600 ms \cite{indefrey2011spatial}. Turn-taking is therefore \textit{predictive}: listeners plan their next turn while the speaker is still speaking \cite{levinson2015timing,garrod2015use}. Multimodal cues including syntax, prosody and gaze support this process \cite{holler2016turn}, enabling speakers to \textit{hold} the floor or to \textit{shift} to another speaker \cite{skantze2021turn}. 

An important goal of human-robot interaction (HRI) is to develop machines capable of real-time conversation \cite{marge2021}. Turn-taking models are central to this objective. Current turn-taking models in consumer systems are \textit{reactive}, deciding whether to speak after a human stops speaking. This results in dialogue that is less spontaneous than human interaction \cite{li2022can,woodruff2003push}. Predictive turn-taking models (PTTMs) aim to overcome these issues \cite{skantze2021turn}. Inspired by human turn-taking, PTTMs are neural networks trained to continually make turn-taking predictions, e.g. the probability of an upcoming shift \cite{skantze2017towards}. Most work on PTTMs has been conducted using corpora of two-party human interaction \cite{skantze2021turn}. 

Almost all PTTMs rely on speech to make predictions; visual information, such as facial expression, is ignored. This may suffice for PTTMs trained on telephone speech \cite{skantze2017towards,li2022can}, but when interlocutors can see one another, might visual cues be useful predictors? The psycholinguistics literature strongly suggests so. \citet{barkhuysen2008interplay} cropped short segments from recordings of questions and asked participants if they belonged to the middle or the end of the question. They were more accurate when presented with both audio and video than in audio-only and video-only conditions. In a recent study \citet{nota2023conversational} found that listeners were faster at recognising questions containing eyebrow frowns. Such studies underline the importance of visual cues in turn-taking and multimodal interaction more generally \cite{holler2019multimodal}.

\paragraph{Aims}

Despite the essential role visual cues play in turn-taking, state-of-the-art PTTMs rely only on speech. It is therefore unknown whether visual cues can improve PTTM performance. We therefore consider the following research questions: 
\begin{enumerate}
    \item Can visual cues improve the performance of predictive turn-taking models? 
    \item If so, which aspects of turn-taking benefit most from the inclusion of visual cues?
\end{enumerate}

\paragraph{Overview}
We introduce multimodal VAP (MM-VAP), a transformer-based PTTM which combines speech with visual features, including facial expression. We show our model outperforms the voice activity projection (VAP) model, a state-of-the-art audio-only PTTM \cite{ekstedt22_interspeech}. 

PTTMs are typically trained using accurate phonetic alignments \cite{ekstedt22_interspeech,li2022can}. We use automatic speech recognition (ASR) to reflect real-word conditions. We re-implement and re-train the VAP model on the Switchboard corpus of telephone speech \cite{godfrey1992switchboard}. We find performance falls slightly due to automatic alignment errors, but the impact is minimal. Next, we train the VAP model using audio from the Candor corpus of videoconferencing interaction \cite{reece2023candor}, again using ASR. We find good performance in distinguishing \textit{shifts} from \textit{holds}. We conduct a facial action unit analysis, which reveals that the next speaker displays enhanced mouth, lip, jaw and chin movements before speaking (i.e. before a shift). These are visual cues which multimodal PTTM could exploit. We therefore combine speech with facial expression, gaze, and head pose in our new multimodal PTTM and demonstrate superior performance over the VAP model (79\% vs 83\% balanced accuracy for hold/shift prediction). The performance increase is most notable in the $F_1$ score for shifts, with a 6-10\% increase in the $F_1$ score. 

In prior work, PTTM performance is reported as a single figure for all holds and shifts. In a more comprehensive analysis, we group holds and shifts by duration of silence between turns. This reveals the strength of our multimodal model across all durations of silence between turns (gaps) and overlapping speech. The performance of both MM-VAP and VAP is worse when there are longer gaps in the Candor corpus, however, MM-VAP outperforms the audio-only model across all durations of gaps considered (83\% vs 79\% for all gaps >0 ms, 78\% vs 75\% for all gaps >750 ms). We discuss our working hypothesis that interlocutors utilise visual cues \textit{when they can see one another}, and they must therefore be included to maintain robust performance. 

Finally, we conduct a detailed ablation study. This reveals that facial action units, which encode facial expression, are the biggest contributors to the increased accuracy over the audio-only turn-taking model. 

We conclude with a discussion of our findings, highlighting the vital importance of non-verbal communication, which researchers in human-robot interaction must consider going forward. We make our code publicly available for future research\footnote{\url{https://github.com/russelsa/mm-vap}}. 

\section{Background}
\label{sec:background}

\paragraph{Turn-taking}
In a conversation, the current speaker either \textit{holds} the turn or \textit{shifts} to another interlocutor \cite{sacks1974simplest}. The time between turns is the floor-transfer offset (FTO), which is positive for a gap and negative for an overlap \cite{heldner2010pauses}. A typical two-party interaction has a mean FTO of +200 ms \cite{stivers2009universals}. There is no single definition of a turn. In Conversation Analysis, a turn is defined by social action (a question, agreement, etc.) \cite{kasper2014conversation}. In predictive turn-taking, a turn is identified from voice activity (VA), which indicates the presence or absence of speech at any moment in time \cite{ekstedt22_interspeech,li2022can}. 

Turn-taking is aided by multimodal cues. The words that we speak as well as how we speak (prosody) play important roles. \citet{bogels2015listeners} found prosody is essential for listeners to correctly determine the end of the turns with multiple completion points e.g. "\textit{are you a student $\backslash$ here $\backslash$ at this university $\backslash$}.". Visual cues include gaze. Speakers look away at the start and back toward the listener at the end of a turn \cite{kendon1967some}. Certain gestures and facial expressions are associated with faster shifts between speakers \cite{trujillo2021visual,nota2023conversational}. 

\paragraph{Predictive turn-taking}
PTTMs continually predict upcoming speaker changes \cite{skantze2021turn}. PTTMs are either RNN \cite{skantze2017towards,li2022can} or transformer-based \cite{ekstedt22_interspeech}. They can be trained to predict the VA in the next 2 seconds \cite{ekstedt22_interspeech}, enabling turn-taking predictions to be made, e.g. predict a shift if the VA probability of the listener exceeds that of the speaker. The start time of the next turn can also be predicted \cite{li2022can}. PTTMs are trained on corpora of human-human interaction and may be deployed to human-robot interaction later. Corpora used to train PTTMs include Switchboard (telephone speech; \citet{godfrey1992switchboard,li2022can,ekstedt22_interspeech}), MapTask (in-person, video unavailable; \citet{anderson1991hcrc,roddy2018multimodal,roddy2018investigating}) and the Mahnob corpus (in-person, with video recordings; \citet{bilakhia2015mahnob,roddy2018multimodal}). Earlier PTTMs use engineered features e.g. part-of-speech tags and the GeMAPs acoustic feature set \cite{skantze2017towards,roddy2018investigating}, though feature engineering has now been replaced by pre-trained feature extractors \cite{ekstedt22_interspeech}.

\paragraph{Multimodal predictive turn-taking}
Almost all PTTMs for two-party interaction rely on speech alone, though a limited number of multimodal models have been reported. \citet{roddy2018multimodal} demonstrated that incorporating gaze vectors alongside speech in a PTTM resulted in a performance improvement on the 11 hour Mahnob corpus \cite{bilakhia2015mahnob} (0.86 vs 0.85 $F_1$ hold/shift prediction). \citet{kurata23_interspeech} found visual features improved the classification of 5 second segments of speech located at the end of turns into hold and shift categories. As the end of the turn must already be known for the model to run, it is therefore not a PTTM, although the findings are promising. 

Onishi et al. \cite{onishi2024multimodal} proposed to extend a recent state-of-the-art audio-only turn-taking model, the VAP model \cite{ekstedt2022much}, to include visual features. Like Roddy et al., they found that the inclusion of visual cues boosted performance however, the model was only tested on between 1.5-2.0 hours of data in 4 languages. As the audio of two speakers was downmixed into a mono audio channel, ground-truth knowledge of the current speaker (the voice activity) is required for inference \cite{onishi2024multimodal}. The model therefore 'knows' exactly when each speaker change occurs. 

The inclusion of the visual modality is therefore under-explored, and it is unclear if visual information can enhance performance at scale. Though beyond the scope of this work, visual features have been considered in PTTMs for multiparty interaction \cite{malikWhoSpeaksNext2020}.

\paragraph{The use of manual alignments}
The use of time-aligned transcriptions to extract the voice activity for training and testing is widespread in the PTTM literature \cite{roddy2018investigating,ekstedt22_interspeech,onishi2024multimodal,inoue-etal-2024-multilingual}. A manual approach limits the amount of data that can be used for training and testing. Automatic speech recognition (ASR) is a promising tool for conversational speech transcription \cite{russell2024automatic}. However, it introduces errors, particularly on aspects of conversational speech such as filled pauses and disfluencies \cite{russell2024automatic}. To the best of our knowledge, the use of ASR transcribed interaction for training and testing PTTMs has not been considered in the literature to date, despite more closely mirroring a real-world deployment and enabling much larger quantities of data to be used for training and testing. 

\section{Methodology} 
\label{sec:method}

\begin{figure}
    \centering
    \includegraphics[width=\columnwidth]{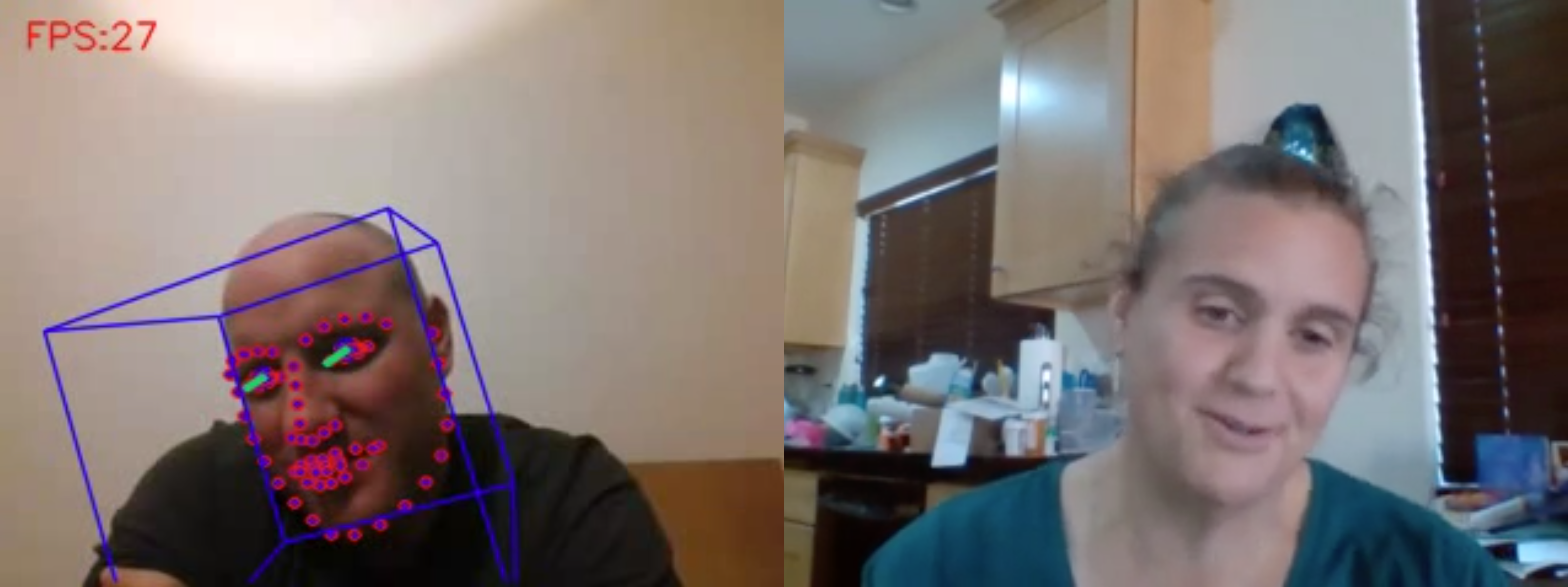}
    \caption{Still from Candor with OpenFace features shown for the left participant}
    \label{fig:candor-still}
\end{figure}

\paragraph{Corpora} We use the Candor corpus \cite{reece2023candor} of 1,656 two-party videoconferencing (VC) interactions to train and validate PTTMs. Each interaction consists of casual conversation in US English e.g. sports teams (mean session length 34 mn). Its large size is ideal for deep learning. Although not in-person, interlocutors can see one another. We are unaware of a suitable in-person corpus of this size. VC provides a front-facing camera angle that is ideal for visual feature extraction (Figure \ref{fig:candor-still}). We use a 710 hour subset of the full 850 hour corpus where visual feature extraction is optimal, detailed further on. We obtain a time-aligned transcription using \citet{speechmatics_asr} ASR previously validated on VC speech \cite{russell2024automatic}. We also use the 260 hour Switchboard corpus of two-party US English interactions \cite{godfrey1992switchboard}. Participants discuss a prescribed topic, e.g. office attire (mean session length 6 min 23 s). Prior PTTMs rely on accurate phonetic transcriptions like those provided with Switchboard \cite{ekstedt22_interspeech}. Using ASR introduces an inevitable alignment error of approximately 480 ms \cite{russell2024automatic}, though this reflects real-world conditions.  
Both corpora contain stereo audio (one channel per speaker), which we downsample to 16 kHz. Candor has 320x240 resolution, 30 fps mp4 video. 

\begin{figure}[!h]
    \centering
    \includegraphics[width=\linewidth]{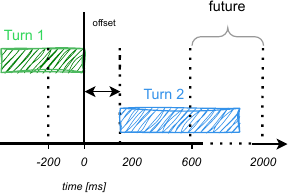}
    \caption{Schematic depicting a shift between speakers}
    \label{fig:shift-schematic}
\end{figure}

\paragraph{Identifying turn-taking events} 
We extract \texttt{shifts} and \texttt{holds} from the transcriptions by identifying silences greater than +250 ms where only one speaker is active 1 second before and after the silence \citet{ekstedt22_interspeech}. If the speaker remains the same, it is a \texttt{hold} and if the speaker changes, it is a \texttt{shift} (see Figure \ref{fig:shift-schematic}). We repeat the above procedure for different FTOs to assess performance across longer and shorter holds and shifts, and overlapping speech. This is a more comprehensive analysis than prior work which aggregates all holds and shifts together \cite{ekstedt22_interspeech}. In Table \ref{tab:corpus-turn-taking-statistics} we show the complete set of holds and shifts in each corpus. Note that the ASR and ground-truth alignments result in different numbers of holds and shifts in Switchboard, due to ASR's inherent alignment error. 

\begin{table}[!ht]
\centering
\caption{Statistics for shifts and holds. Floor-transfer offset (FTO) is the time between the end of the previous turn and the start of the next turn (negative overlap).}
\label{tab:corpus-turn-taking-statistics}
\resizebox{0.95\linewidth}{!}{%
\begin{tabular}{lrrrrrr}
\toprule
  \multicolumn{1}{l}{\textbf{Corpus}}  &
  \multicolumn{1}{l}{\textbf{Minium}}  &
  \multicolumn{1}{l}{\textbf{\# Shifts}}  &
  \multicolumn{1}{l}{\textbf{\# Holds}}  &
  \multicolumn{1}{c}{\textbf{Shifts}}  &
  \multicolumn{1}{c}{\textbf{\# Shifts}} &
  \multicolumn{1}{c}{\textbf{\# Holds}} \\
  \multicolumn{1}{l}{}  &
  \multicolumn{1}{c}{\textbf{FTO [ms]}} &
  \multicolumn{1}{l}{} &
  \multicolumn{1}{l}{} &
  \multicolumn{1}{l}{\textbf{proportion}} &
  \multicolumn{1}{c}{\textbf{per minute}} &
  \multicolumn{1}{c}{\textbf{per minute}} \\ \midrule
\textit{Candor}   & -250 & 23515  & 62483  & 0.38 & 0.42 & 1.11  \\
\textit{(ASR)}         & 0    & 115975 & 414666 & 0.28 & 2.05 & 7.34  \\
  710 hours       & 250  & 83158  & 206830 & 0.40 & 1.47 & 3.66  \\
         & 500  & 53360  & 115475 & 0.46 & 0.94 & 2.04  \\
         & 750  & 32831  & 65804  & 0.50 & 0.58 & 1.16  \\
         & 1000 & 19699  & 35697  & 0.55 & 0.35 & 0.63  \\
         & 1250 & 11438  & 19831  & 0.58 & 0.20 & 0.35  \\
         & 1500 & 6148   & 10250  & 0.60 & 0.11 & 0.18  \\ \cline{1-7}
\textit{Switchboard}   & -250 & 6599   & 9419   & 0.70 & 0.42 & 0.61  \\
\textit{(ground-truth}          & 0    & 26883  & 150407 & 0.18 & 1.73 & 9.67  \\
 \textit{alignment)}        & 250  & 16267  & 74510  & 0.22 & 1.05 & 4.79  \\
  260 hours       & 500  & 9909   & 43793  & 0.23 & 0.64 & 2.82  \\
         & 750  & 6087   & 27348  & 0.22 & 0.39 & 1.76  \\
         & 1000 & 3893   & 14180  & 0.27 & 0.25 & 0.91  \\
         & 1250 & 2407   & 6958   & 0.35 & 0.15 & 0.45  \\
         & 1500 & 1385   & 3348   & 0.41 & 0.09 & 0.22  \\ \cline{1-7}
\textit{Switchboard} & -250 & 10595  & 7761   & 1.37 & 0.68 & 0.50  \\
\textit{(ASR }        & 0    & 34886  & 210386 & 0.17 & 2.24 & 13.53 \\
\textit{    alignment)}         & 250  & 20097  & 143182 & 0.14 & 1.29 & 9.21  \\
  260 hours       & 500  & 10302  & 74666  & 0.14 & 0.66 & 4.80  \\
         & 750  & 5269   & 33299  & 0.16 & 0.34 & 2.14  \\
         & 1000 & 2702   & 14952  & 0.18 & 0.17 & 0.96  \\
         & 1250 & 1383   & 6599   & 0.21 & 0.09 & 0.42  \\
         & 1500 & 663    & 2786   & 0.24 & 0.04 & 0.18 \\ \bottomrule

\end{tabular}%
}
\end{table}

\paragraph{Visual feature extraction}
We use OpenFace \cite{baltruvsaitis2016openface} to extract visual features. Facial action units (FAUs, 17 in total) numerically describe facial movements e.g. jaw drop on a scale from $0.0$ to $5.0$. There is one gaze vector per eye which is a $3$ dimensional unit vector. We also extract head position ($X, Y, Z$ in mm) and head rotation (roll, pitch and yaw in radians). We select 15 facial landmarks located on the brow, jaw, nose and lips ($x$ and $y$ in pixels) and a confidence score. In total, there is one $60$ dimensional vector per frame. We scale all features to 0, 1 by max min scaling at participant level. Histograms of eye gaze and head pose are unimodal with different modes per participant, reflecting the differing setup of participant devices. We therefore zero mean the head pose and eye gaze vectors at participant level. We show a sample visualisation of OpenFace features in Figure \ref{fig:candor-still}. Head pose is depicted as a blue cube centred on the head pointing in the estimated direction of head pose. Eye gaze vectors are depicted in green. Facial landmarks are shown in red, though we only use a subset of 15 in this work. It is not straightforward to visualise FAUs so these are omitted. 

OpenFace fails to completely track participants in 238 of the sessions. We investigate and find failure is due to various issues e.g. in one session, a participant gets up and leaves to fetch something. In the overwhelming majority of the corpus (1,418 sessions, 710 hours) OpenFace runs without issue. We manually check tracking performance on a subset and observe good performance. We therefore conduct all our work on these sessions. 

For analysis, we compute maximum FAU intensity 200 ms before shifts and holds where the FTO > + 250 ms, a common time frame in PTTM evaluation \cite{ekstedt22_interspeech,roddy2018investigating}. We select an equal number of 200 ms random periods of silence and speech located far (1 second) away from the start or end of a turn. We then compare the median FAU intensity.

\begin{figure}[!h]
    \centering
    \includegraphics[width=0.8\linewidth]{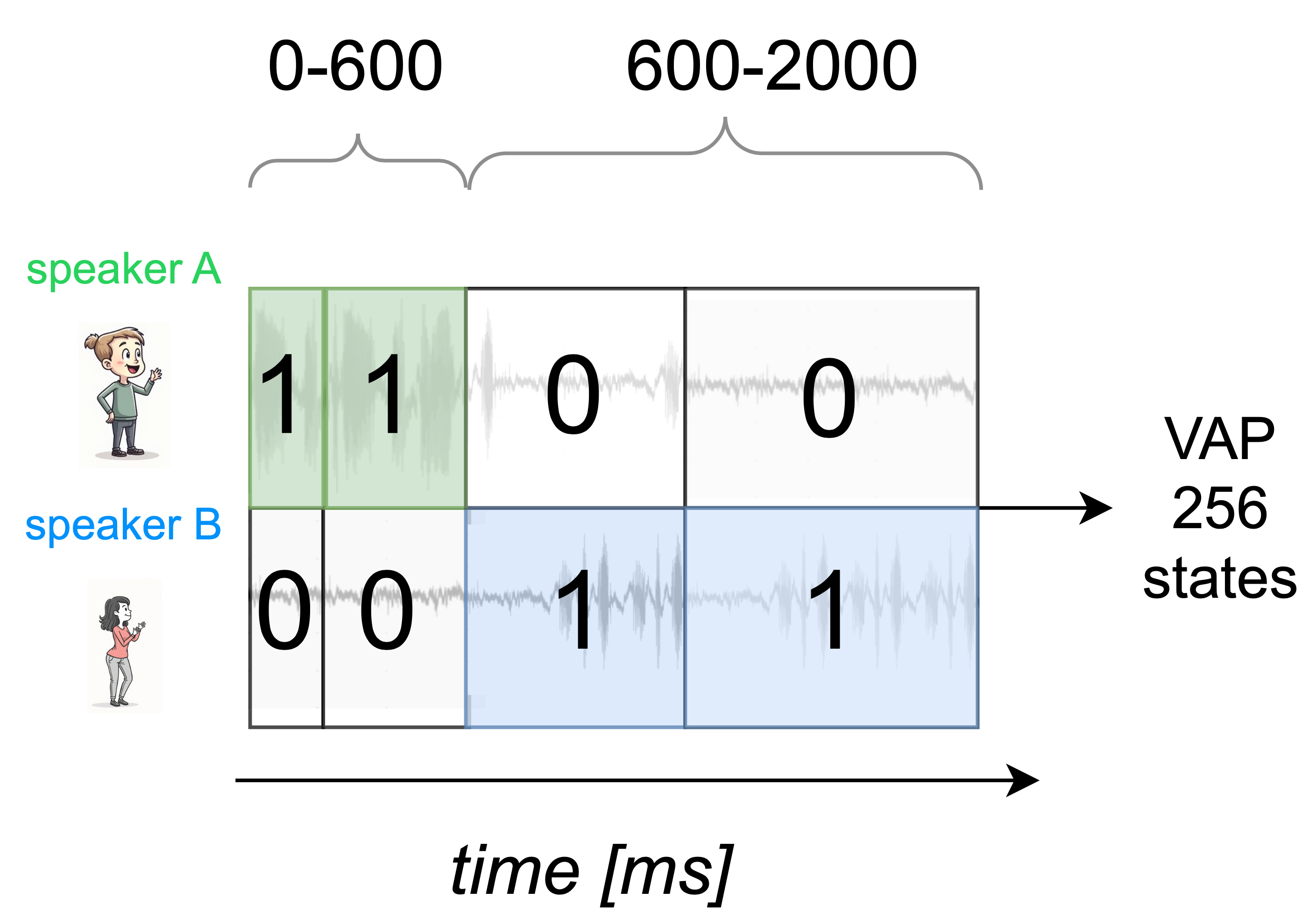}
    \caption{The VAP training objective introduced by \citet{ekstedt22_interspeech} which captures speaking activity in the next 2 seconds in a two-party interaction. 
    }
    \label{fig:vap-objective}
\end{figure}

\paragraph{Audio-only turn-taking model}
We establish the performance of the state-of-the-art voice activity projection (VAP) model \cite{ekstedt22_interspeech}, an audio-only turn-taking model which has been used in several studies e.g. \cite{ekstedt2022much,ekstedt23_interspeech,inoue-etal-2024-multilingual}. The VAP model is a transformer \cite{vaswani2017attention} based neural network trained to predict who will speak in the next 2 seconds: the \textit{training objective}, shown in Figure \ref{fig:vap-objective}. At each point in time there are four bins per speaker representing the next 2 seconds. The bins span the next 0-200, 200-600, 600-1200 and 1200-2000 milliseconds. If 50\% of frames within a bin contain speech, we set the bin label to 1, otherwise the bin is labelled 0. This gives $2^8=256$ possible VAP states. The model has 5.8 M parameters, and full details are in \citet{inoue-etal-2024-multilingual,ekstedt2022much}.

\paragraph{Multimodal turn-taking model (ours)}

\begin{figure}[!h]
    \centering
    \includegraphics[width=0.45\linewidth]{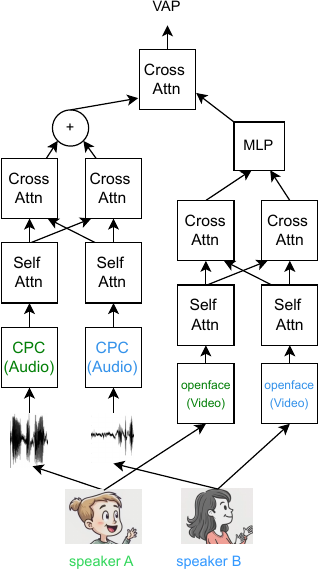}
    \caption{Schematic of our transformer-based multimodal predictive turn-taking model (late fusion version), incorporating audio and video from both speakers. }
    \label{fig:multimodal-tt-model-schematic}
\end{figure}

Like the VAP model, our model, multimodal VAP (MM-VAP, Figure \ref{fig:multimodal-tt-model-schematic}) consists of self- and cross-attention blocks. A self-attention block is $N$ stacked transformer decoder layers, where query ($q$), key ($k$) and value ($v$) are identical: 

\vspace{-1.5em}
\begin{equation*}
\resizebox{\linewidth}{!}{%
    $\text{\textsc{Self-Attn}}(x) = \text{\textsc{Transformer}}\left( q = x,\ k = x,\ v = x \right)$%
}
\end{equation*}
\vspace{-1em}

A transformer layer learns attention or the 'compatibility' of output with the query via the key and value \cite{vaswani2017attention}. In self-attention blocks, as $q=k=v$, the model learns temporal patterns in the input e.g. audio or video. In cross-attention blocks, we stack $N$ layers with two inputs $x_1$ and $x_2$. In each layer, we compute two transformer layers with shared weights: 

\vspace{-1.5em}
\begin{equation*}
\resizebox{\columnwidth}{!}{%
    $\text{\textsc{Cross-Attn}}(x_1, x_2) = \sigma 
    \brkbinom{
        \text{\textsc{Transformer}}\left( q = x_1,\, k = x_2,\, v = x_2 \right) + 
    }{
        \text{\textsc{Transformer}}\left( q = x_2,\, k = x_1,\, v = x_1 \right)
    }$
}
\end{equation*}
\vspace{-1em}

where $\sigma ( \cdot )$ is a layer normalisation operation followed by a GeLU activation. In these layers, the model learns temporal patterns between $x_1$ and $x_2$ e.g. from audio to video.

The model in Figure \ref{fig:multimodal-tt-model-schematic} is a \textit{late fusion} model as the audio and video modalities are combined just prior to the output \cite{baltruvsaitis2018multimodal}. The initial cross-attention layers learn patterns between each speaker's audio and video separately. The final cross-attention block learns temporal patterns between the video and audio modalities. We also design an early fusion model where audio and video from each participant is fed into a cross-attention layer just after the feature extraction. Both versions have 8.7 M parameters. As each speaker's visual and non-verbal signals are modelled separately and combined with cross-attention, the voice activity signal is not required at inference. 

Like the VAP model, we use a pre-trained audio feature extractor \cite{riviere2020unsupervised} yielding a 256 dimensional feature vector at 50 Hz. We freeze the extractor layers in training. We upsample visual features from 30 to 50 Hz to match using linear interpolation. Within the model, we use single hidden layer multilayer perceptrons (MLPs) to project the 60 dimensional vector to 256 dimensions. Self-attention blocks consist of 3 stacked self-attention layers and 1 cross-attention block. We apply a causal masking to ensure that the model can only make predictions from past audio and video frames. We use layer normalisation \cite{lei2016layer} and the GeLU activation function \cite{hendrycks2016gaussian} throughout. The model outputs a 256 dimensional vector via the softmax function, learning a probability distribution over all 256 VAP states with a cross-entropy loss (Figure \ref{fig:vap-objective}). 

\paragraph{Training}

We train all models as follows. We withhold 5\% of sessions for testing. We conduct a 5-fold cross-validation on the remaining sessions (80\% training, 20\% validation). We segment audio and video into 20 second segments with a 2 second overlap and randomly shuffle all segments. We use the cross-entropy loss function to train the model output (256 dimensions) to learn VAP labels (Figure \ref{fig:vap-objective}). Based on an initial hyperparameter sweep we set the batch size to 16 and the learning rate to 0.005. We train the model for all 5 folds with an Nvidia RTX 6000 GPU for 10 epochs. The GPU hours are: VAP Switchboard 21 hrs, VAP Candor 21 hrs, early fusion 110 hrs, late fusion 113 hrs. 

\paragraph{Evaluation}

We evaluate trained models as follows. At each hold or shift in the validation set, we sum model probabilities in a 200 ms window. The window is located either before the end of a turn, the start of an overlap, or during mutual silence between speaking turns. We sum the shift probability, defined as the marginal probability of all VAP states where the non-active speaker is 1 in both bins in the 600-2000 ms period (Figure \ref{fig:vap-objective}). Like the original VAP paper \cite{ekstedt22_interspeech} we only consider the latter half of the VAP objective for hold/shift evaluation. We find that predictions are less reliable in the first 0-600 ms. We then set a threshold: if the summed probability is greater than this threshold, a shift is predicted. We choose thresholds which maximise the weighted $F_1$ and balanced accuracy scores on the validation set. Finally, we report performance on the unseen test sessions with these thresholds. We compare $F_1$ and balanced accuracy with a dummy baseline model which always outputs a hold. 

We report performance with the $F_1$ and balanced accuracy. The $F_1$ is \cite{powers2020evaluation}:

\begin{equation}
    F_1 =  \frac{2 \times \# \textsc{TP}}{2 \times \# \textsc{TP} + \# \textsc{FP} + \# \textsc{FN}}
\end{equation}
\vspace{-0.6em}

where TP is the number of true positives, FP false positives and FN false negatives. We compute the $F_1$ score by arbitrarily assigning a positive 1 label to a shift and a negative 0 to a hold ($F_1$ shift). We then re-compute with a positive 1 = hold and a 0 = shift ($F_1$ hold) and report the weighed $F_1$:

\begin{equation}
    F_{1,weighted} = \rho_{s} F_1 \textsc{shift}  + \rho_{h} F_1 \textsc{hold} 
\end{equation}

where $\rho_{s}$ and $\rho_{h}$ are the proportion of shifts and holds. The weighted $F_1$ accounts for the presence of more holds than shifts (Table \ref{tab:corpus-turn-taking-statistics}). We also report balanced accuracy \cite{balanced_acc}: 

\begin{equation}
    \textsc{Bal. accuracy} = \frac{1}{2} \left( \frac{\#TP}{\#P} + \frac{\#TN}{\#N} \right)
\end{equation}

where TP and TF are the number of true positives and false negatives, and P and N are the numbers of positives (shifts) and negatives (holds). We compare $F_1$ and balanced accuracy of models on the common test set with the paired t-test \cite{Ross2017}. We compare facial action units with the non-parametric Mann-Whitney U (MWU) test \cite{MWU_test}. 

\paragraph{Ablation study}
We conduct an ablation study by training 4 different versions of the MM-VAP model. We train and validate using the same procedure, but each of which receives audio and a subset of the visual features as input. We divide the visual features into gaze, head pose, facial action unit and facial landmarks groups (6, 6, 17, 30 dimensions receptively, excluding the scalar confidence score). 

\section{Results}
\label{sec:results}

\subsection{Audio-only turn-taking}

We begin by assessing the performance of the audio-only VAP predictive turn-taking model (PTTM). We first consider hold/shift prediction during periods of silence greater than 250 ms, as in \citet{ekstedt22_interspeech}. We report the 5-fold average performance in Table \ref{tab:audio-only-ttm}. 

\begin{table}[!h]
\centering
\caption{Average shift/hold prediction performance of the VAP model on Candor (CND, videoconference) and Switchboard (SWB, telephone) corpora evaluated during silence between turns (FTO > +250 ms). }
\label{tab:audio-only-ttm}
\resizebox{\columnwidth}{!}{%
\begin{tabular}{llcccc}
\toprule
  \multicolumn{1}{l}{\textbf{Corpus}} &
  \textbf{Alignment} &
  \multicolumn{1}{c}{$\mathbf{F_1}$ \textbf{(Weighted)}} &
  \multicolumn{1}{c}{$\mathbf{F_1}$ \textbf{(Hold)}} &
  \multicolumn{1}{c}{$\mathbf{F_1}$ \textbf{(Shift)}} &
  \multicolumn{1}{c}{\textbf{\makecell{Accuracy \\ (Balanced \%)}}} 
  \\ \hline
  \textit{SWB} & ground-truth  & 0.82 & 0.89 & 0.47 & 67 \\
  \textit{SWB} & ASR & 0.81 & 0.89 & 0.45 & 65 \\
  \textit{CND} & ASR & 0.83 & 0.89 & 0.71 & 79 \\
  \bottomrule
\end{tabular}%
}
\end{table}

All models trained on Switchboard outperform a baseline model which always predicts hold (weighted $F_1$: $0.74$ for SWB ASR and $0.70$ for Candor, $p<0.01$). The Switchboard ground-truth alignment results are comparable with those of \citet{ekstedt22_interspeech}, verifying our re-implementation. The $F_1$ and balanced accuracy scores are slightly higher using the ground-truth alignment, reflecting the ASR alignment error. In the Candor corpus, using audio-only cues, the VAP model performs well above the baseline and the balanced accuracy is higher than Switchboard. There is a higher $F_1$ shift and a similar $F_1$ hold and weighted $F_1$. 

\subsection{Visual feature analysis}

\begin{figure}[!h]
    \centering
    \includegraphics[width=\columnwidth]{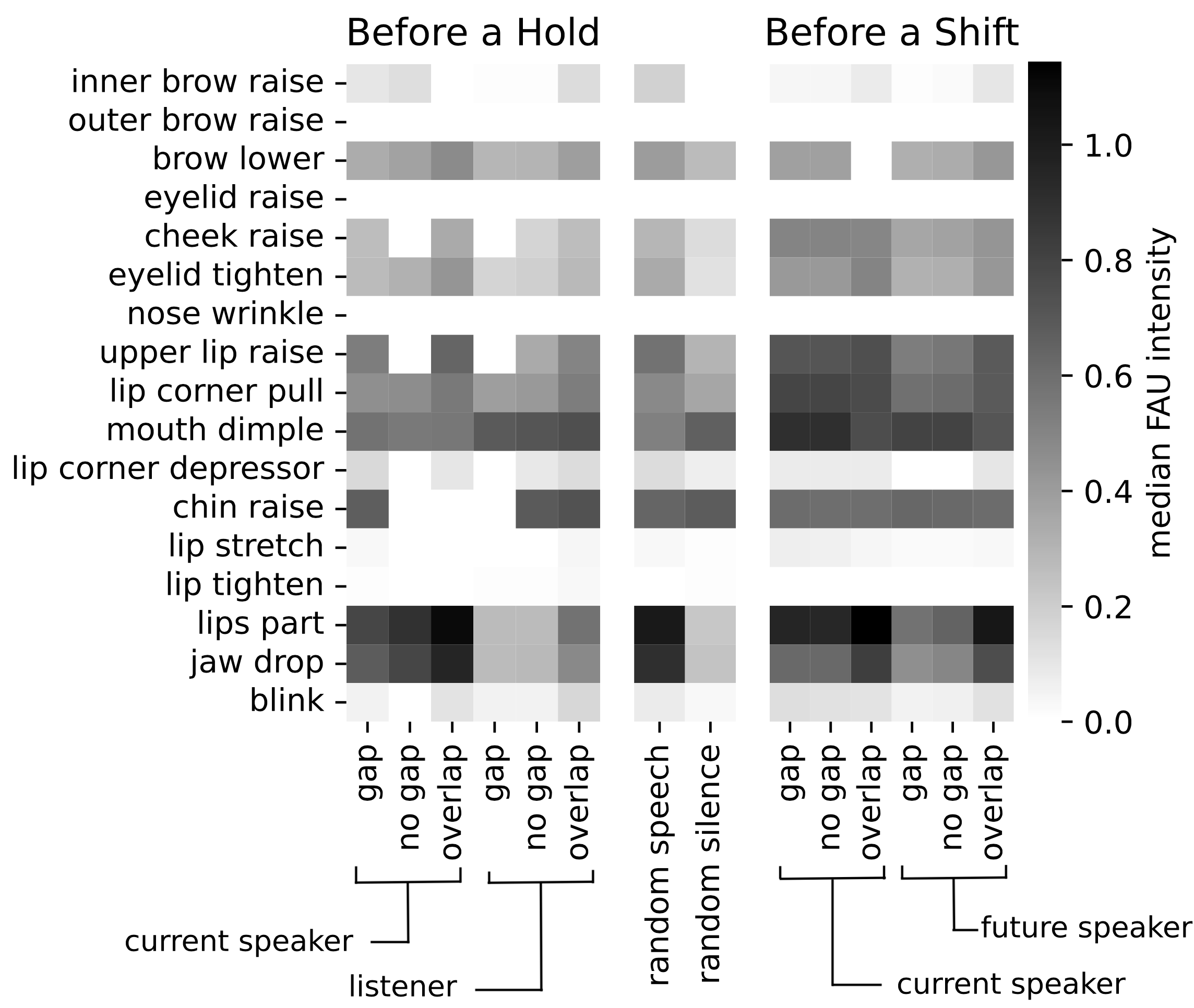}
    \caption{Median FAU intensity in Candor during random speech and silence, and before holds and shifts.}
    \label{fig:fau-intensity-heatmap}
\end{figure}

As turn-taking literature details the importance of visual cues (Section \ref{sec:intro}), we investigate facial expression in Candor interactions. In Figure \ref{fig:fau-intensity-heatmap} we show the median peak FAU intensity at key turn-taking events (method in Section \ref{sec:method}). We compare with random speech and random silence with the MWU test and omit comparisons where $p>0.05$. We exclude outer brow raiser, upper eyelid raiser and nose wrinkler as $p>0.05$ in all comparisons. 

As a sanity check, we observe that when an interlocutor speaks, there is more cheek, eyelid, brow, lip and jaw activity than when they do not speak (\textit{random speech} and \textit{random silence}). At the end of a turn, the FAU activity of the current speaker largely resembles random speech. During a hold, the listener remains silent and FAU activity resembles random silence (\textit{listener} vs \textit{random silence}). We observe subtle differences in the FAU activity of shifts when comparing cheek, lip, and jaw FAUs of \textit{future speaker} with \textit{random silence}. This suggests the presence of a distinct facial expression partially resembling speech ahead of an upcoming transition. This could be exploited in a predictive turn-taking model. We repeated the analysis for eye gaze and head pose. These did not reveal meaningful patterns, so we omit heat maps. These features may still be useful in a neural network which captures patterns across much long periods of time. 

\subsection{Multimodal turn-taking}

We continue by investigating the performance of our proposed early and late fusion turn-taking models (MM-VAP). We train on the Candor corpus (Section \ref{sec:method}). For completeness, we also train a VAP model only using video features. This model is identical to the original VAP model with the removal of the audio feature extractor and a dimensionality of 60, reflecting the dimensionality of the visual feature vector. 


\begin{table}[!h]
\centering
\caption{Average shift/hold prediction across 5 folds on the Candor corpus (best in bold). a = audio only, v = video-only, and a+v = multimodal; e = early fusion, l = late fusion. Percentage change relative to the audio-only model is provided as $\uparrow$ / $\downarrow$, with '-' indicating $p > 0.05$.}
\label{tab:multimodal-model-accuracy}
\resizebox{\columnwidth}{!}{%
\begin{tabular}{llrrrrrrrr}
\toprule
\textbf{Evaluation point} &
\textbf{Model} &
\multicolumn{2}{c}{$\mathbf{F_1}$ \textbf{(Weighted)}} &
\multicolumn{2}{c}{$\mathbf{F_1}$ \textbf{(Hold)}} &
\multicolumn{2}{c}{$\mathbf{F_1}$ \textbf{(Shift)}} &
\multicolumn{2}{c}{\textbf{\makecell{Accuracy \\ (Balanced \%)}}} \\
\midrule

\multicolumn{1}{r}{\textit{during mutual silence}}  & a         & 0.83 &        & 0.88 &        & 0.70 &        & 79 &        \\
\multicolumn{1}{r}{\textit{(FTO > +250 ms)}} & v         & 0.72 & $\downarrow$14\% & 0.82 & $\downarrow$8\% & 0.47 & $\downarrow$33\% & 68 & $\downarrow$15\% \\
& a+v (e)   & 0.84 & -    & 0.90 & - & 0.74 & $\uparrow$5\%    & 82 & $\uparrow$3\% \\
& \textbf{a+v (l)} & \textbf{0.86} & $\uparrow$3\% & \textbf{0.90} & $\uparrow$2\% & \textbf{0.74} & $\uparrow$6\% & \textbf{83} & $\uparrow$4\% \\
\midrule

\multicolumn{1}{r}{\textit{before end of turn}} & a         & 0.81 &        & 0.87 &        & 0.67 &        & 76 &        \\
\multicolumn{1}{r}{\textit{(FTO > +250 ms)}} & v         & 0.70 & $\downarrow$13\% & 0.80 & $\downarrow$8\% & 0.45 & $\downarrow$33\% & 66 & $\downarrow$14\% \\
& a+v (e)   & 0.83 & $\uparrow$2\%    & 0.88 & -    & 0.69 & $\uparrow$4\%    & 79 & $\uparrow$3\% \\
& \textbf{a+v (l)} & \textbf{0.83} & $\uparrow$3\% & \textbf{0.89} & $\uparrow$2\% & \textbf{0.70} & $\uparrow$6\% & \textbf{80} & $\uparrow$4\% \\
\midrule

\multicolumn{1}{r}{\textit{before end of turn}} & a         & 0.86 &        & 0.91 &        & 0.66 &        & 77 &        \\
\multicolumn{1}{r}{\textit{(FTO > 0 ms)}} & v         & 0.78 & $\downarrow$10\% & 0.87 & $\downarrow$5\% & 0.43 & $\downarrow$35\% & 70 & $\downarrow$10\% \\
& a+v (e)   & 0.87 & $\uparrow$2\%    & 0.92 & -    & 0.69 & $\uparrow$6\%    & 80 & $\uparrow$3\% \\
& \textbf{a+v (l)} & \textbf{0.87} & $\uparrow$2\% & \textbf{0.92} & - & \textbf{0.71} & $\uparrow$6\% & \textbf{83} & $\uparrow$4\% \\
\midrule

\multicolumn{1}{r}{\textit{before overlap}} & a         & 0.78 &        & 0.85 &        & 0.57 &        & 70 &        \\
\multicolumn{1}{r}{\textit{(FTO < -250 ms)}} & v         & 0.72 & $\downarrow$8\%  & 0.82 & $\downarrow$3\%  & 0.42 & $\downarrow$25\% & 64 & $\downarrow$8\% \\
& a+v (e)   & 0.79 & $\uparrow$3\%    & 0.87 & $\uparrow$2\%    & 0.60 & $\uparrow$7\%    & 72 & $\uparrow$3\% \\
& \textbf{a+v (l)} & \textbf{0.80} & $\uparrow$4\% & \textbf{0.87} & $\uparrow$2\% & \textbf{0.62} & $\uparrow$10\% & \textbf{74} & $\uparrow$5\% \\

\bottomrule
\end{tabular}%
}
\end{table}

\paragraph{Multimodal vs. audio-only models}
We report the average performance over 5 folds of the Candor corpus in Table \ref{tab:multimodal-model-accuracy}. As before, we consider model performance 200 ms before a shift/hold (FTO > 250 ms). Our multimodal models significantly outperform the audio-only and video-only models ($p<0.01$). The late fusion model shows a $3$\% relative increase in the weighted $F_1$ score and a $6$\% relative increase in balanced accuracy over the audio-only model. The video-only model has the worst overall performance with a 33\% relative reduction in the $F_1$ shift score. 

\paragraph{Expanding the analysis}
Next, we remove the requirement for a minimum period of silence between turns, increasing the number of shifts and holds considered (FTO > 0 vs FTO > 250 ms in Table \ref{tab:corpus-turn-taking-statistics}). We also move the evaluation window to 200 ms before the end of a turn, ensuring the next speaker has not started to speak during evaluation. This hence captures the capacity of the model to predict upcoming shifts while the previous speaker is still speaking, akin to human turn-taking (Section \ref{sec:intro}). For this more comprehensive set of speaker transitions, our multimodal models outperforms both video-only and audio-only models. Best performance is achieved with the late fusion model, with a $6\%$ relative increase in the $F_1$ shift score and a $4\%$ relative increase in balanced accuracy (Table \ref{tab:multimodal-model-accuracy}). We also evaluate hold/shift prediction before overlapping speech. Note that for the purposes of this evaluation, we exclude overlapping speech where there is no change in speaker 1 second after the overlap has concluded. This ensures that brief periods of overlap are excluded (i.e. backchannels). We find a 6\% increase in balanced accuracy and a $10\%$ increase in the $F_1$ shift scores for the late-fusion MM-VAP model (Table \ref{tab:multimodal-model-accuracy}). 

\paragraph{Performance during longer transitions}

\begin{figure}[!htpb]
    \centering
    \includegraphics[width=\linewidth]{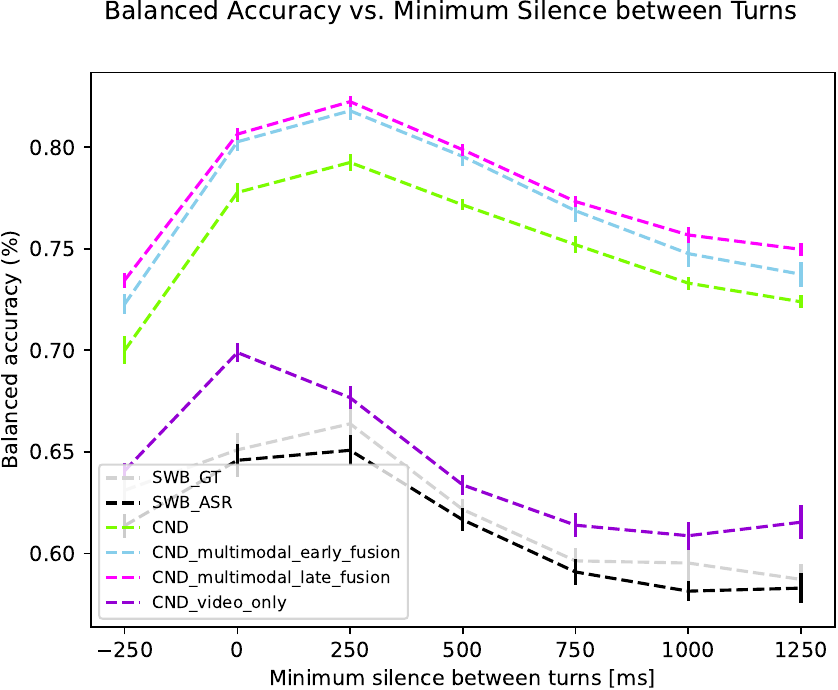}
    \caption{Balanced accuracy of models averaged over 5 folds, $\pm$ standard error in the mean, grouped by a minimum period of silence between turns; i.e. the FTO.}
    \label{fig:balanced-accuracy-minimum-silence-period}
\end{figure}

Our multimodal models are the best-performing overall. We assess if this performance benefit is uniform across all types of transitions. We group subsets of holds and shifts by varying the minimum FTO from 0 to 1500 ms in 250 ms increments (Table \ref{tab:corpus-turn-taking-statistics}). We then compute model performance on each subset. We use the balanced accuracy because, unlike the $F1$ weighted score, it equally weights holds and shifts. This is important as proportions of holds/shifts by group (Table \ref{tab:corpus-turn-taking-statistics}). We plot the mean balanced accuracy over 5 folds in Figure \ref{fig:balanced-accuracy-minimum-silence-period}. 

The balanced accuracies in Tables \ref{tab:audio-only-ttm} and \ref{tab:multimodal-model-accuracy} correspond with the 0 and 250 ms points in Figure \ref{fig:balanced-accuracy-minimum-silence-period}. The gap in performance of Switchboard and Candor is consistent with our prior finding of a lower $F_1$ shift score on Switchboard (Table \ref{tab:audio-only-ttm}). Performance falls for both audio-only VAP and multimodal MM-VAP models as the duration of silence between speaking turns increases, indicating that these turns are more challenging to predict. However, we find that the performance of the MM-VAP models is significantly better than the audio-only VAP models across all FTOs ($p<0.01$ in all cases). Thus, the inclusion of visual information leads to improved performance across the complete range of holds and shifts in the corpus. 

\paragraph{Ablation study}
Finally, we conduct an ablation study, training the late fusion version of the MM-VAP model on audio and gaze, head pose, facial landmarks and facial action unit subsets. We report the 5-fold average performance in Table \ref{tab:ablation_study}. 

\begin{table}[!h]
\centering
\caption{Ablation study results, 5-fold average. a=audio, v=video, a+v (l) = multimodal, late fusion. Percentage change relative to the a+v (l) model which uses all visual features is shown as $\uparrow$ / $\downarrow$, with '-' indicating $p > 0.05$.}
\label{tab:ablation_study}
\resizebox{\columnwidth}{!}{%
\begin{tabular}{llrrrrrrrr}
\toprule
\textbf{Evaluation Point} & \textbf{Model} &
\multicolumn{2}{c}{$\mathbf{F_1}$ \textbf{(Weighted)}} &
\multicolumn{2}{c}{$\mathbf{F_1}$ \textbf{(Hold)}} &
\multicolumn{2}{c}{$\mathbf{F_1}$ \textbf{(Shift)}} &
\multicolumn{2}{c}{\textbf{\makecell{ Accuracy \\  (Balanced \%)}}} \\
\midrule

\multicolumn{1}{r}{\textit{during mutual silence}} & a + v (l)   & 0.86 &        & 0.90 &        & 0.74 &        & 83 &        \\
\multicolumn{1}{r}{\textit{(FTO > +250 ms)}} 
    & a + gaze  & 0.74 & $\downarrow$14\% & 0.83 & $\downarrow$8\%  & 0.51 & $\downarrow$31\% & 67 & $\downarrow$20\% \\
    & a + pose  & 0.84 & $\downarrow$2\%  & 0.89 & -  & 0.72 & $\downarrow$3\%  & 80 & $\downarrow$3\% \\
    & a + lmks  & 0.79 & $\downarrow$8\%  & 0.86 & $\downarrow$4\%  & 0.62 & $\downarrow$17\% & 74 & $\downarrow$11\% \\
    & a + faus  & 0.85 & -  & 0.89 & -  & 0.73 & $\downarrow$2\%  & 81 & $\downarrow$2\% \\

\midrule
\multicolumn{1}{r}{\textit{before end of turn}} & a + v (l)   & 0.83 &        & 0.88 &        & 0.74 &        & 80 &        \\
\multicolumn{1}{r}{\textit{(FTO > 250 ms)}} 
    & a + gaze  & 0.72 & $\downarrow$13\% & 0.82 & $\downarrow$8\%  & 0.48 & $\downarrow$35\% & 65 & $\downarrow$19\% \\
    & a + pose  & 0.82 & -  & 0.87 & -  & 0.68 & $\downarrow$8\%  & 78 & $\downarrow$2\% \\
    & a + lmks  & 0.78 & $\downarrow$7\%  & 0.85 & $\downarrow$4\%  & 0.59 & $\downarrow$20\% & 72 & $\downarrow$10\% \\
    & a + faus  & 0.83 & -  & 0.88 & -  & 0.69 & $\downarrow$6\%  & 79 & $\downarrow$2\% \\

\midrule
\multicolumn{1}{r}{\textit{before end of turn}} & a + v (l)   & 0.87 &        & 0.92 &        & 0.71 &        & 83 &        \\
\multicolumn{1}{r}{\textit{(FTO > 0 ms)}} 
    & a + gaze  & 0.78 & $\downarrow$10\% & 0.88 & $\downarrow$5\%  & 0.43 & $\downarrow$39\% & 65 & $\downarrow$20\% \\
    & a + pose  & 0.87 & -  & 0.92 & -  & 0.67 & $\downarrow$3\%  & 79 & $\downarrow$2\% \\
    & a + lmks  & 0.83 & $\downarrow$5\%  & 0.90 & $\downarrow$3\%  & 0.57 & $\downarrow$18\% & 73 & $\downarrow$10\% \\
    & a + faus  & 0.87 & -                & 0.92 & -                & 0.68 & $\downarrow$2\%  & 80 & $\downarrow$4\% \\

\midrule
\multicolumn{1}{r}{\textit{overlap}} & a + v (l)   & 0.80 &        & 0.87 &        & 0.62 &        & 74 &        \\
\multicolumn{1}{r}{\textit{(FTO < 250 ms)}} 
    & a + gaze  & 0.68 & $\downarrow$15\% & 0.81 & $\downarrow$7\%  & 0.35 & $\downarrow$44\% & 58 & $\downarrow$22\% \\
    & a + pose  & 0.79 & $\downarrow$2\%  & 0.86 & -  & 0.58 & $\downarrow$6\%  & 71 & $\downarrow$3\% \\
    & a + lmks  & 0.74 & $\downarrow$8\%  & 0.84 & $\downarrow$4\%  & 0.49 & $\downarrow$22\% & 66 & $\downarrow$11\% \\
    & a + faus  & 0.79 & $\downarrow$2\%  & 0.86 & -  & 0.59 & $\downarrow$4\%  & 72 & $\downarrow$2\% \\

\bottomrule
\end{tabular}%
}
\end{table}

The best-performing model is the model which includes all visual features (a + v (l), Table \ref{tab:ablation_study}). Considering the models trained on the four subsets of the visual features, we note reduced performance, which is most notable when considering the $F_1$ shift score. We compare the weighted $F1$ score, and find that the gaze and landmark trained models perform significantly worse than the model trained on all visual features in all cases ($p<0.01$ when comparing a+v (l) with gaze and landmarks, Table \ref{tab:ablation_study}). There is no significant difference in the performance of the facial action unit and head pose trained models on certain metrics (e.g. $p=0.09$, before end of turn FTO > 0ms Table \ref{tab:ablation_study}). However, on the $F1$ shift score, the facial action unit trained model achieves the best performance of all models trained on the reduced visual feature sets ($p<0.05$). The best-performing model trained on a subset of visual features is therefore the facial action unit model. 

\section{Discussion}
\label{sec:discussion}

Our re-implementation of the audio-only VAP PTTM \cite{ekstedt22_interspeech,inoue-etal-2024-multilingual} performed well on the Candor videoconferencing (VC) corpus. However, our new multimodal PTTM, MM-VAP - which uses speech along with facial action units, landmarks, gaze, and head pose - significantly outperformed the VAP model. The performance increase was most notable on the $F_1$ shift metric, with a 6-10\% relative increase in performance (Table \ref{tab:audio-only-ttm}). Returning to our first research question, our findings show that visual cues improve PTTM performance. This echoes the psycholinguistics literature underlining the importance of visual cues in turn-taking (Section \ref{sec:intro}).

We also investigated the use of ASR, as to date PTTM models have used manual alignments. We found performance dropped slightly but remained broadly similar, supporting the use of ASR in PTTM development (Table \ref{tab:audio-only-ttm}). 

\paragraph{How does visual information help?}

Our second research question concerned the aspects of turn-taking which benefit from visual information. As prior work considered all holds and shifts together, we conducted a more fine-grained analysis. This revealed that shifts, or transitions between speakers, benefit the most from visual information. Our working hypothesis is that non-verbal behaviours provide vital turn-taking cues when interlocutors can see one another, and our model benefits from this. Notwithstanding slight differences in the corpora (session length, topic, see Section \ref{sec:method}), the most notable difference is that in Candor interlocutors can see one another, whereas in Switchboard, they cannot. Thus, as our results show, worse performance is achieved when audio alone is used to model turn-taking in Candor. Our multimodal model which incorporates visual cues improved performance, and we demonstrated that this is consistent across the complete range of holds and shifts in the corpus. A detailed analysis using Conversation Analysis methodology should be conducted to uncover the exact role visual cues play here, however we found some evidence of enhanced lip jaw and cheek movement before speaking (Figure \ref{fig:fau-intensity-heatmap}). This indicates the presence of visual cues from the listener prior to the onset of speech (a shift). This is consistent with our finding that visual cues most benefited turn-taking by improving the performance on shifts ($F_1$ shift, Table \ref{tab:multimodal-model-accuracy}). 

\paragraph{Visual signalling}
As the role of visual cues in turn-taking is well-supported by the psycholinguistics literature (Sections \ref{sec:intro} and \ref{sec:background}), we believe visual cues such as gaze aversion are exploited in our multimodal model. Future work is needed to establish the exact role of visual cues and their impact on model performance, but the literature outlines why visual cues are particularly important during speaker transitions. For instance, the more complex a question is, the longer the response \cite{strombergsson2013timing} e.g. open-ended versus yes/no questions \cite{walczyk2003cognitive}. Extended silences are also associated with gaze aversion in interaction \cite{walczyk2003cognitive} and hence these may rely more on non-verbal cues. 

Finally, we conducted a thorough ablation study, which showed that facial action units are the biggest contributors to model performance. The ablation study demonstrated that not all features contributed to model performance in isolation, most notably gaze (Table \ref{tab:ablation_study}). This is despite the role of gaze in turn-taking being well-supported in the turn-taking literature \cite{kendon1967some}. The videoconferencing setup of the corpus \cite{reece2023candor} might be impacting performance here, as videoconferencing is known to impact gaze \cite{sellenRemoteConversationsEffects1995}. Alternatively, gaze extraction is challenging and can be impacted by factors such as lighting levels \cite{zhang2021eye}, which are uncontrolled in the Candor corpus. However, we did find that when combined together, the model which incorporated all visual features performed the best (Table \ref{tab:ablation_study}), suggesting the model does leverage information from all of visual features. 

\paragraph{Future work}
MM-VAP has shown promising results, but it can be improved. OpenFace could be replaced by a more powerful learnable visual front-end, e.g. from the audio-visual speech recognition literature \cite{fenghour2021deep}. The model should be updated to handle audio and video at different sample rates (Section \ref{sec:method}). 

\section{Conclusion}
\label{sec:conclusion}

We presented the first comprehensive analysis of multimodal predictive turn-taking, introducing MM-VAP a new multimodal model which uses speech along with visual features. We found a strong improvement in performance above the audio-only state-of-the-art in videoconferencing speech. We therefore encourage researchers to incorporate multimodal cues into models for predictive turn-taking and multimodal interaction more generally. We make all code publicly available. 

\section{Acknowledgements} 
This research was conducted with the financial support of Science Foundation Ireland under Grant Agreement No. 13/RC/2106\_P2 at the ADAPT SFI Research Centre at Trinity College Dublin. Speechmatics provided academic access to their ASR platform.

\section{Limitations}
\label{sec:limitations}

Our paper has a number of inevitable limitations, which we discuss below. 

\paragraph{Data limitations} We compared telephone and videoconferencing (VC) interaction and did not include any in-person corpora. Videoconferencing interaction has a number of key differences to in-person interaction, including limited use of body language and difficulties in knowing when to speak \cite{sellenRemoteConversationsEffects1995,isaacsWhatVideoCan,oconaillConversationsVideoConferences1993,bolandZoomDisruptsRhythm2021}. Nevertheless, we have shown that including visual features improves performance in PTTMs trained on VC data. We expect that this would also be the case in in-person settings, where interlocutors are not hindered in their use of non-verbal communication by technology, though this should be confirmed. A comparison of a model trained on a corpus of in-person interaction would be useful. However, this is not straightforward due to the lack of availability of a suitable public dataset. Available corpora of dyadic interaction are not large, e.g. the 11-hour Mahnob mimicry corpus \cite{bilakhia2015mahnob}. Furthermore, varying camera angles can complicate visual feature extraction, unlike in VC where the angle is front-facing. The use of ambient microphones in in-person settings introduces issues such as background noise and speaker diarisation. These issues are resolved through the advanced audio processing algorithms in modern VC platforms. 

\paragraph{Feature extraction limitations} A further limitation is the use of OpenFace \cite{baltruvsaitis2016openface} for visual feature extraction. OpenFace has the benefit of being easily interpretable by humans, through the extraction of high-level features like head pose, but this may not be the most useful representation to use in a neural network. However, the model does benefit from these features as shown by our results. As we have suggested for future work, the audio-visual speech recognition literature is a good starting point for visual front-end architectures \cite{fenghour2021deep,lipresding}. OpenFace is also not robust as it failed on a minority of sessions. This is due to issues beyond our control, which are an inherent part of data captured \textit{in-the-wild}. Excluding these sessions leaves 710 hours; more than sufficient for deep learning. Future front-ends should be made more robust to missing data arising from issues such as participants leaving the frame. We did not assess how OpenFace tracking may also be hampered by variable lighting conditions, glasses, facial hair, etc. but we observed that tracking was good in sessions we verified manually. Again, these artefacts will be unavoidable in data captured in naturalistic interaction. The Candor and Switchboard corpora, though captured in naturalistic settings, are free from any specific background noises, and the audio quality is good. In less controlled settings, the ASR transcription could be of lower quality than the one used here. This is due to the impact of background noises, echo, poor microphone quality, all of which degrade ASR performance \cite{agrawal2019modulation,alharbiAutomaticSpeechRecognition2021}.

\paragraph{Model limitations}
Our multimodal model itself has its own limitations. It is unable to handle video and audio features at different frame rates (30 Hz / fps for the video and 50 Hz for the audio after feature extraction). We resolved this by upsampling the visual features in time to 50 Hz (i.e. a factor of 1.67). We applied causal masking to ensure the model could not use future information to make predictions, though the upsampling does introduce a slight future 'bleed' as present frames are modified by the next frame through interpolation. A future version of the model could overcome this by ensuring that visual and acoustic features are handled at different temporal rates as done with an RNN in \cite{roddy2018multimodal}. The acoustic features are processed by passing them through a pre-trained feature extractor \cite{riviere2020unsupervised}, whereas the visual features are high-level descriptors, e.g. angles in radians. This issue could be resolved in a future iteration of the model, replacing the OpenFace front-end, as discussed. Nevertheless, the model introduced in this paper shows considerable performance improvements over the state-of-the-art audio-only approach.

\bibliography{custom}

\begin{thebibliography}{55}
\providecommand{\natexlab}[1]{#1}

\bibitem[{Agrawal and Ganapathy(2019)}]{agrawal2019modulation}
Purvi Agrawal and Sriram Ganapathy. 2019.
\newblock Modulation filter learning using deep variational networks for robust speech recognition.
\newblock \emph{IEEE journal of selected topics in signal processing}, 13(2):244--253.

\bibitem[{Alharbi et~al.(2021)Alharbi, Alrazgan, Alrashed, Alnomasi, Almojel, Alharbi, Alharbi, Alturki, Alshehri, and Almojil}]{alharbiAutomaticSpeechRecognition2021}
Sadeen Alharbi, Muna Alrazgan, Alanoud Alrashed, Turkiayh Alnomasi, Raghad Almojel, Rimah Alharbi, Saja Alharbi, Sahar Alturki, Fatimah Alshehri, and Maha Almojil. 2021.
\newblock \href {https://doi.org/10.1109/ACCESS.2021.3112535} {Automatic {Speech} {Recognition}: {Systematic} {Literature} {Review}}.
\newblock \emph{IEEE Access}, 9:131858--131876.

\bibitem[{Anderson et~al.(1991)Anderson, Bader, Bard, Boyle, Doherty, Garrod, Isard, Kowtko, McAllister, Miller et~al.}]{anderson1991hcrc}
Anne~H Anderson, Miles Bader, Ellen~Gurman Bard, Elizabeth Boyle, Gwyneth Doherty, Simon Garrod, Stephen Isard, Jacqueline Kowtko, Jan McAllister, Jim Miller, et~al. 1991.
\newblock The hcrc map task corpus.
\newblock \emph{Language and speech}, 34(4):351--366.

\bibitem[{Baltru{\v{s}}aitis et~al.(2018)Baltru{\v{s}}aitis, Ahuja, and Morency}]{baltruvsaitis2018multimodal}
Tadas Baltru{\v{s}}aitis, Chaitanya Ahuja, and Louis-Philippe Morency. 2018.
\newblock Multimodal machine learning: A survey and taxonomy.
\newblock \emph{IEEE transactions on pattern analysis and machine intelligence}, 41(2):423--443.

\bibitem[{Baltru{\v{s}}aitis et~al.(2016)Baltru{\v{s}}aitis, Robinson, and Morency}]{baltruvsaitis2016openface}
Tadas Baltru{\v{s}}aitis, Peter Robinson, and Louis-Philippe Morency. 2016.
\newblock Openface: an open source facial behavior analysis toolkit.
\newblock In \emph{2016 IEEE winter conference on applications of computer vision (WACV)}, pages 1--10. IEEE.

\bibitem[{Barkhuysen et~al.(2008)Barkhuysen, Krahmer, and Swerts}]{barkhuysen2008interplay}
Pashiera Barkhuysen, Emiel Krahmer, and Marc Swerts. 2008.
\newblock The interplay between the auditory and visual modality for end-of-utterance detection.
\newblock \emph{The journal of the Acoustical Society of America}, 123(1):354--365.

\bibitem[{Bilakhia et~al.(2015)Bilakhia, Petridis, Nijholt, and Pantic}]{bilakhia2015mahnob}
Sanjay Bilakhia, Stavros Petridis, Anton Nijholt, and Maja Pantic. 2015.
\newblock The mahnob mimicry database: A database of naturalistic human interactions.
\newblock \emph{Pattern recognition letters}, 66:52--61.

\bibitem[{B{\"o}gels and Torreira(2015)}]{bogels2015listeners}
Sara B{\"o}gels and Francisco Torreira. 2015.
\newblock Listeners use intonational phrase boundaries to project turn ends in spoken interaction.
\newblock \emph{Journal of Phonetics}, 52:46--57.

\bibitem[{Boland et~al.(2021)Boland, Fonseca, Mermelstein, and Williamson}]{bolandZoomDisruptsRhythm2021}
Julie~E Boland, Pedro Fonseca, Ilana Mermelstein, and Myles Williamson. 2021.
\newblock \href {https://doi.org/10.1037/xge0001150} {Zoom disrupts the rhythm of conversation.}
\newblock \emph{Journal of Experimental Psychology: General}, 151(6):1272–1282.

\bibitem[{Brodersen et~al.(2010)Brodersen, Ong, Stephan, and Buhmann}]{balanced_acc}
Kay~Henning Brodersen, Cheng~Soon Ong, Klaas~Enno Stephan, and Joachim~M. Buhmann. 2010.
\newblock \href {https://doi.org/10.1109/ICPR.2010.764} {The balanced accuracy and its posterior distribution}.
\newblock In \emph{2010 20th International Conference on Pattern Recognition}, pages 3121--3124.

\bibitem[{Ekstedt and Skantze(2022{\natexlab{a}})}]{ekstedt2022much}
Erik Ekstedt and Gabriel Skantze. 2022{\natexlab{a}}.
\newblock How much does prosody help turn-taking? investigations using voice activity projection models.
\newblock In \emph{Proceedings of the 23rd Annual Meeting of the Special Interest Group on Discourse and Dialogue}, pages 541--551.

\bibitem[{Ekstedt and Skantze(2022{\natexlab{b}})}]{ekstedt22_interspeech}
Erik Ekstedt and Gabriel Skantze. 2022{\natexlab{b}}.
\newblock \href {https://doi.org/10.21437/Interspeech.2022-10955} {Voice activity projection: Self-supervised learning of turn-taking events}.
\newblock In \emph{Interspeech 2022}, pages 5190--5194.

\bibitem[{Ekstedt et~al.(2023)Ekstedt, Wang, Éva Székely, Gustafson, and Skantze}]{ekstedt23_interspeech}
Erik Ekstedt, Siyang Wang, Éva Székely, Joakim Gustafson, and Gabriel Skantze. 2023.
\newblock \href {https://doi.org/10.21437/Interspeech.2023-2064} {Automatic evaluation of turn-taking cues in conversational speech synthesis}.
\newblock In \emph{INTERSPEECH 2023}, pages 5481--5485.

\bibitem[{Fenghour et~al.(2021)Fenghour, Chen, Guo, Li, and Xiao}]{fenghour2021deep}
Souheil Fenghour, Daqing Chen, Kun Guo, Bo~Li, and Perry Xiao. 2021.
\newblock Deep learning-based automated lip-reading: A survey.
\newblock \emph{IEEE Access}, 9:121184--121205.

\bibitem[{Garrod and Pickering(2015)}]{garrod2015use}
Simon Garrod and Martin~J Pickering. 2015.
\newblock The use of content and timing to predict turn transitions.
\newblock \emph{Frontiers in psychology}, 6(751):1--12.

\bibitem[{Godfrey et~al.(1992)Godfrey, Holliman, and McDaniel}]{godfrey1992switchboard}
John~J Godfrey, Edward~C Holliman, and Jane McDaniel. 1992.
\newblock Switchboard: Telephone speech corpus for research and development.
\newblock In \emph{Acoustics, speech, and signal processing, ieee international conference on}, volume~1, pages 517--520. IEEE Computer Society.

\bibitem[{Heldner and Edlund(2010)}]{heldner2010pauses}
Mattias Heldner and Jens Edlund. 2010.
\newblock Pauses, gaps and overlaps in conversations.
\newblock \emph{Journal of Phonetics}, 38(4):555--568.

\bibitem[{Hendrycks and Gimpel(2016)}]{hendrycks2016gaussian}
Dan Hendrycks and Kevin Gimpel. 2016.
\newblock Gaussian error linear units (gelus).
\newblock \emph{arXiv preprint arXiv:1606.08415}.

\bibitem[{Holler et~al.(2016)Holler, Kendrick, Casillas, and Levinson}]{holler2016turn}
Judith Holler, Kobin~H Kendrick, Marisa Casillas, and Stephen~C Levinson. 2016.
\newblock \emph{Turn-taking in human communicative interaction}.
\newblock Frontiers Media SA.

\bibitem[{Holler and Levinson(2019)}]{holler2019multimodal}
Judith Holler and Stephen~C Levinson. 2019.
\newblock Multimodal language processing in human communication.
\newblock \emph{Trends in Cognitive Sciences}, 23(8):639--652.

\bibitem[{Indefrey(2011)}]{indefrey2011spatial}
Peter Indefrey. 2011.
\newblock The spatial and temporal signatures of word production components: a critical update.
\newblock \emph{Frontiers in psychology}, 2:255.

\bibitem[{Inoue et~al.(2024)Inoue, Jiang, Ekstedt, Kawahara, and Skantze}]{inoue-etal-2024-multilingual}
Koji Inoue, Bing{'}er Jiang, Erik Ekstedt, Tatsuya Kawahara, and Gabriel Skantze. 2024.
\newblock \href {https://aclanthology.org/2024.lrec-main.1036} {Multilingual turn-taking prediction using voice activity projection}.
\newblock In \emph{Proceedings of the 2024 Joint International Conference on Computational Linguistics, Language Resources and Evaluation (LREC-COLING 2024)}, pages 11873--11883, Torino, Italia. ELRA and ICCL.

\bibitem[{Isaacs and Tang(1993)}]{isaacsWhatVideoCan}
Ellen~A Isaacs and John~C Tang. 1993.
\newblock What video can and can't do for collaboration: a case study.
\newblock In \emph{Proceedings of the first ACM International Conference on Multimedia}, pages 199--206. ACM Press.

\bibitem[{Ivanko et~al.(2023)Ivanko, Ryumin, and Karpov}]{lipresding}
Denis Ivanko, Dmitry Ryumin, and Alexey Karpov. 2023.
\newblock \href {https://doi.org/10.3390/math11122665} {A review of recent advances on deep learning methods for audio-visual speech recognition}.
\newblock \emph{Mathematics}, 11(12).

\bibitem[{Kasper and Wagner(2014)}]{kasper2014conversation}
Gabriele Kasper and Johannes Wagner. 2014.
\newblock Conversation analysis in applied linguistics.
\newblock \emph{Annual Review of Applied Linguistics}, 34:171--212.

\bibitem[{Kendon(1967)}]{kendon1967some}
Adam Kendon. 1967.
\newblock Some functions of gaze-direction in social interaction.
\newblock \emph{Acta psychologica}, 26:22--63.

\bibitem[{Kurata et~al.(2023)Kurata, Saeki, Fujie, and Matsuyama}]{kurata23_interspeech}
Fuma Kurata, Mao Saeki, Shinya Fujie, and Yoichi Matsuyama. 2023.
\newblock \href {https://doi.org/10.21437/Interspeech.2023-578} {Multimodal turn-taking model using visual cues for end-of-utterance prediction in spoken dialogue systems}.
\newblock In \emph{INTERSPEECH 2023}, pages 2658--2662.

\bibitem[{Lei~Ba et~al.(2016)Lei~Ba, Kiros, and Hinton}]{lei2016layer}
Jimmy Lei~Ba, Jamie~Ryan Kiros, and Geoffrey~E Hinton. 2016.
\newblock Layer normalization.
\newblock \emph{ArXiv e-prints}, pages arXiv--1607.

\bibitem[{Levinson and Torreira(2015)}]{levinson2015timing}
Stephen~C Levinson and Francisco Torreira. 2015.
\newblock Timing in turn-taking and its implications for processing models of language.
\newblock \emph{Frontiers in psychology}, 6:731.

\bibitem[{Li et~al.(2022)Li, Paranjape, and Manning}]{li2022can}
Siyan Li, Ashwin Paranjape, and Christopher~D Manning. 2022.
\newblock When can i speak? predicting initiation points for spoken dialogue agents.
\newblock \emph{arXiv preprint arXiv:2208.03812}.

\bibitem[{Malik et~al.(2020)Malik, Saunier, Funakoshi, and Pauchet}]{malikWhoSpeaksNext2020}
Usman Malik, Julien Saunier, Kotaro Funakoshi, and Alexandre Pauchet. 2020.
\newblock \href {https://doi.org/10.1109/ICTAI50040.2020.00062} {Who {Speaks} {Next}? {Turn} {Change} and {Next} {Speaker} {Prediction} in {Multimodal} {Multiparty} {Interaction}}.
\newblock In \emph{2020 {IEEE} 32nd {International} {Conference} on {Tools} with {Artificial} {Intelligence} ({ICTAI})}, pages 349--354, Baltimore, MD, USA. IEEE.

\bibitem[{Mann and Whitney(1947)}]{MWU_test}
H.~B. Mann and D.~R. Whitney. 1947.
\newblock \href {https://doi.org/10.1214/aoms/1177730491} {{On a Test of Whether one of Two Random Variables is Stochastically Larger than the Other}}.
\newblock \emph{The Annals of Mathematical Statistics}, 18(1):50 -- 60.

\bibitem[{Marge et~al.(2022)Marge, Espy-Wilson, Ward, Alwan, Artzi, Bansal, Blankenship, Chai, Daumé, Dey, Harper, Howard, Kennington, Kruijff-Korbayová, Manocha, Matuszek, Mead, Mooney, Moore, Ostendorf, Pon-Barry, Rudnicky, Scheutz, Amant, Sun, Tellex, Traum, and Yu}]{marge2021}
Matthew Marge, Carol Espy-Wilson, Nigel~G. Ward, Abeer Alwan, Yoav Artzi, Mohit Bansal, Gil Blankenship, Joyce Chai, Hal Daumé, Debadeepta Dey, Mary Harper, Thomas Howard, Casey Kennington, Ivana Kruijff-Korbayová, Dinesh Manocha, Cynthia Matuszek, Ross Mead, Raymond Mooney, Roger~K. Moore, Mari Ostendorf, Heather Pon-Barry, Alexander~I. Rudnicky, Matthias Scheutz, Robert~St. Amant, Tong Sun, Stefanie Tellex, David Traum, and Zhou Yu. 2022.
\newblock \href {https://doi.org/10.1016/j.csl.2021.101255} {Spoken language interaction with robots: Recommendations for future research}.
\newblock \emph{Computer Speech and Language}, 71:101255.

\bibitem[{Nota et~al.(2023)Nota, Trujillo, and Holler}]{nota2023conversational}
Naomi Nota, James~P Trujillo, and Judith Holler. 2023.
\newblock Conversational eyebrow frowns facilitate question identification: An online study using virtual avatars.
\newblock \emph{Cognitive Science}, 47(12):e13392.

\bibitem[{O'Conaill et~al.(1993)O'Conaill, Whittaker, and Wilbur}]{oconaillConversationsVideoConferences1993}
Brid O'Conaill, Steve Whittaker, and Sylvia Wilbur. 1993.
\newblock \href {https://doi.org/10.1207/s15327051hci0804_4} {Conversations {Over} {Video} {Conferences}: {An} {Evaluation} of the {Spoken} {Aspects} of {Video}-{Mediated} {Communication}}.
\newblock \emph{Human–Computer Interaction}, 8(4):389--428.

\bibitem[{Onishi et~al.(2024)Onishi, Tanaka, and Nakamura}]{onishi2024multimodal}
Kazuyo Onishi, Hiroki Tanaka, and Satoshi Nakamura. 2024.
\newblock Multimodal voice activity projection for turn-taking and effects on speaker adaptation.
\newblock \emph{IEICE Transactions on Information and Systems}.

\bibitem[{Powers(2020)}]{powers2020evaluation}
David~MW Powers. 2020.
\newblock Evaluation: from precision, recall and f-measure to roc, informedness, markedness and correlation.
\newblock \emph{arXiv preprint arXiv:2010.16061}.

\bibitem[{Reece et~al.(2023)Reece, Cooney, Bull, Chung, Dawson, Fitzpatrick, Glazer, Knox, Liebscher, and Marin}]{reece2023candor}
Andrew Reece, Gus Cooney, Peter Bull, Christine Chung, Bryn Dawson, Casey Fitzpatrick, Tamara Glazer, Dean Knox, Alex Liebscher, and Sebastian Marin. 2023.
\newblock The candor corpus: Insights from a large multimodal dataset of naturalistic conversation.
\newblock \emph{Science Advances}, 9(13):eadf3197.

\bibitem[{Riviere et~al.(2020)Riviere, Joulin, Mazar{\'e}, and Dupoux}]{riviere2020unsupervised}
Morgane Riviere, Armand Joulin, Pierre-Emmanuel Mazar{\'e}, and Emmanuel Dupoux. 2020.
\newblock Unsupervised pretraining transfers well across languages.
\newblock In \emph{ICASSP 2020-2020 IEEE International Conference on Acoustics, Speech and Signal Processing (ICASSP)}, pages 7414--7418. IEEE.

\bibitem[{Roddy et~al.(2018{\natexlab{a}})Roddy, Skantze, and Harte}]{roddy2018investigating}
Matthew Roddy, Gabriel Skantze, and Naomi Harte. 2018{\natexlab{a}}.
\newblock Investigating speech features for continuous turn-taking prediction using lstms.
\newblock \emph{arXiv preprint arXiv:1806.11461}.

\bibitem[{Roddy et~al.(2018{\natexlab{b}})Roddy, Skantze, and Harte}]{roddy2018multimodal}
Matthew Roddy, Gabriel Skantze, and Naomi Harte. 2018{\natexlab{b}}.
\newblock Multimodal continuous turn-taking prediction using multiscale rnns.
\newblock In \emph{Proceedings of the 20th ACM International Conference on Multimodal Interaction}, pages 186--190.

\bibitem[{Ross and Willson(2017)}]{Ross2017}
Amanda Ross and Victor~L. Willson. 2017.
\newblock \href {https://doi.org/10.1007/978-94-6351-086-8_3} {\emph{Independent Samples T-Test}}, pages 13--16.
\newblock SensePublishers, Rotterdam.

\bibitem[{Russell et~al.(2024)Russell, Gessinger, Krason, Vigliocco, and Harte}]{russell2024automatic}
Sam~O’Connor Russell, Iona Gessinger, Anna Krason, Gabriella Vigliocco, and Naomi Harte. 2024.
\newblock What automatic speech recognition can and cannot do for conversational speech transcription.
\newblock \emph{Research Methods in Applied Linguistics}, 3(3):100163.

\bibitem[{Sacks et~al.(1974)Sacks, Schegloff, and Jefferson}]{sacks1974simplest}
Harvey Sacks, Emanuel~A Schegloff, and Gail Jefferson. 1974.
\newblock A simplest systematics for the organization of turn-taking for conversation.
\newblock \emph{language}, 50(4):696--735.

\bibitem[{Sellen(1995)}]{sellenRemoteConversationsEffects1995}
Abigail Sellen. 1995.
\newblock \href {https://doi.org/10.1207/s15327051hci1004_2} {Remote {Conversations}: {The} {Effects} of {Mediating} {Talk} {With} {Technology}}.
\newblock \emph{Human-Computer Interaction}, 10(4):401--444.

\bibitem[{Skantze(2017)}]{skantze2017towards}
Gabriel Skantze. 2017.
\newblock Towards a general, continuous model of turn-taking in spoken dialogue using lstm recurrent neural networks.
\newblock In \emph{Proceedings of the 18th Annual SIGdial Meeting on Discourse and Dialogue}, pages 220--230.

\bibitem[{Skantze(2021)}]{skantze2021turn}
Gabriel Skantze. 2021.
\newblock Turn-taking in conversational systems and human-robot interaction: a review.
\newblock \emph{Computer Speech and Language}, 67:101178.

\bibitem[{{Speechmatics, Ltd.}(2024)}]{speechmatics_asr}
{Speechmatics, Ltd.} 2024.
\newblock \href {https://www.speechmatics.com/company/articles-and-news/introducing-ursa-the-worlds-most-accurate-speech-to-text} {{Speechmatics ASR}}.
\newblock [Online; accessed Dec 2024].

\bibitem[{Stivers et~al.(2009)Stivers, Enfield, Brown, Englert, Hayashi, Heinemann, Hoymann, Rossano, De~Ruiter, Yoon et~al.}]{stivers2009universals}
Tanya Stivers, Nicholas~J Enfield, Penelope Brown, Christina Englert, Makoto Hayashi, Trine Heinemann, Gertie Hoymann, Federico Rossano, Jan~Peter De~Ruiter, Kyung-Eun Yoon, et~al. 2009.
\newblock Universals and cultural variation in turn-taking in conversation.
\newblock \emph{Proceedings of the National Academy of Sciences}, 106(26):10587--10592.

\bibitem[{Str{\"o}mbergsson et~al.(2013)Str{\"o}mbergsson, Hjalmarsson, Edlund, and House}]{strombergsson2013timing}
Sofia Str{\"o}mbergsson, Anna Hjalmarsson, Jens Edlund, and David House. 2013.
\newblock Timing responses to questions in dialogue.
\newblock In \emph{Interspeech}, volume 2013, pages 2584--2588.

\bibitem[{Trujillo et~al.(2021)Trujillo, Levinson, and Holler}]{trujillo2021visual}
James~P Trujillo, Stephen~C Levinson, and Judith Holler. 2021.
\newblock Visual information in computer-mediated interaction matters: Investigating the association between the availability of gesture and turn transition timing in conversation.
\newblock In \emph{Human-Computer Interaction. Design and User Experience Case Studies: Thematic Area, HCI 2021, Held as Part of the 23rd HCI International Conference, HCII 2021, Virtual Event, July 24--29, 2021, Proceedings, Part III 23}, pages 643--657. Springer.

\bibitem[{Vaswani(2017)}]{vaswani2017attention}
A~Vaswani. 2017.
\newblock s.
\newblock \emph{Advances in Neural Information Processing Systems}.

\bibitem[{Walczyk et~al.(2003)Walczyk, Roper, Seemann, and Humphrey}]{walczyk2003cognitive}
Jeffrey~J Walczyk, Karen~S Roper, Eric Seemann, and Angela~M Humphrey. 2003.
\newblock Cognitive mechanisms underlying lying to questions: Response time as a cue to deception.
\newblock \emph{Applied Cognitive Psychology: The Official Journal of the Society for Applied Research in Memory and Cognition}, 17(7):755--774.

\bibitem[{Woodruff and Aoki(2003)}]{woodruff2003push}
Allison Woodruff and Paul~M Aoki. 2003.
\newblock How push-to-talk makes talk less pushy.
\newblock In \emph{Proceedings of the 2003 ACM International Conference on Supporting Group Work}, pages 170--179.

\bibitem[{Zhang et~al.(2021)Zhang, Park, and Maria~Feit}]{zhang2021eye}
Xucong Zhang, Seonwook Park, and Anna Maria~Feit. 2021.
\newblock Eye gaze estimation and its applications.
\newblock \emph{Artificial Intelligence for Human Computer Interaction: A Modern Approach}, pages 99--130.

\end{thebibliography}
\end{document}